\documentclass[11pt, a4paper]{article}
\usepackage[utf8]{inputenc}
\usepackage{lmodern}
\usepackage{geometry}
\usepackage{amsmath}
\usepackage{graphicx}
\usepackage{hyperref}
\usepackage{natbib}

\title{\textbf{\textit{In Silico} Modeling of the RAMPHO Buffer: Dissociating Informational and Energetic Masking via Phonetic Entropy in Deep Neural Networks}}
\author{Stefan Bleeck \\ \textit{Institute of Sound and Vibration Research (ISVR), University of Southampton}}
\date{May 2026}

\begin{document}

\maketitle

\begin{abstract}
The fundamental challenge of listening in multi-talker environments is a cognitive bottleneck, defined by the Ease of Language Understanding (ELU) model as a failure within the RAMPHO episodic buffer. Current deep neural networks for speech enhancement optimize purely for physical acoustics, failing to account for the cognitive penalty of informational masking. Here, we present an \textit{in silico} simulation of the RAMPHO buffer using the frame-by-frame phonetic entropy of a self-supervised acoustic model (wav2vec 2.0). By contrasting a semantically intact distractor with a phase-decorrelated distractor (the Concentration Shield) across a signal-to-noise ratio (SNR) sweep, we successfully dissociate the cognitive penalty of informational distraction from the physical penalty of energetic decay. The simulation reveals a cognitive-acoustic Pareto optimization problem: destroying a distractor's semantic payload provides a release from informational masking at high SNRs, but fundamentally degrades temporal glimpsing cues at low SNRs.
\end{abstract}

\section{Introduction and Theoretical Background}

\textbf{The Cognitive Bottleneck in Speech Perception}\\
The fundamental challenge of listening in multi-talker environments---the cocktail party problem---is not merely an acoustic limitation; it is a cognitive bottleneck. Traditional audiometry and signal processing often treat the auditory system as a broken microphone, focusing exclusively on restoring peripheral audibility. However, human hearing is an integrated information processor. According to the Ease of Language Understanding (ELU) model \citep{ronnberg2013}, speech perception under optimal conditions relies on rapid, implicit processing within the RAMPHO (Rapid Automatic Multimodal Phonological) episodic buffer. This buffer seamlessly maps incoming acoustic features to sub-lexical representations with minimal cognitive overhead. When signal degradation or semantic competition prevents this automatic binding, the implicit system fails. The listener is then forced to explicitly recruit finite, top-down working memory resources to repair the stream, resulting in measurable listening effort and rapid cognitive exhaustion \citep{ronnberg2013, pichora2016}.

\vspace{1em}
\noindent \textbf{Energetic vs. Informational Masking}\\
This RAMPHO failure is driven by two distinct mechanisms. Energetic Masking (EM) occurs at the auditory periphery when the physical energy of a distractor physically obliterates the target signal \citep{moore2007}. In contrast, Informational Masking (IM) is a central, cognitive interference. When a background noise is highly intelligible---such as a competing talker---it carries a clear linguistic payload that involuntarily hijacks the listener's semantic processing networks \citep{kidd2023}. The cognitive cost of IM often exceeds that of EM because the brain must expend explicit executive control to suppress the recognizable semantic track.

Furthermore, human listeners naturally combat masking by exploiting temporal fluctuations in the background noise. Through a process of ``acoustic glimpsing,'' the auditory system stitches together momentary target cues that survive in the temporal dips of a fluctuating masker \citep{cooke2006}. If a masker is steady-state, these temporal windows slam shut, drastically increasing the energetic penalty.

\vspace{1em}
\noindent \textbf{The Methodological Gap in Machine Hearing}\\
Despite decades of psychophysical characterization, modern engineering approaches remain deeply decoupled from these biological realities. Current objective speech metrics, such as the Speech Intelligibility Index (SII) or Perceptual Evaluation of Speech Quality (PESQ), function as idealized peripheries; they evaluate acoustic fidelity but are entirely blind to the cognitive penalty of semantic distraction.

Similarly, state-of-the-art Deep Neural Networks (DNNs) utilized in modern hearing technologies achieve unprecedented Signal-to-Noise Ratio (SNR) enhancements by optimizing strictly for physical acoustics. However, these architectures process audio with near-infinite context and perfect working memory, lacking the biological constraints of the human mind. Consequently, an aggressive noise-reduction algorithm might mathematically improve the SNR of a background talker, inadvertently clarifying its linguistic payload. This effectively unmasks the competing talker, replacing a manageable energetic load with a devastating informational load.

To engineer hearing technologies that genuinely protect cognitive resources rather than merely amplifying sound, we must transition from measuring missing acoustics to mapping the exact computational tipping points of the human brain. We require an \textit{in silico} model of the RAMPHO buffer capable of dissociating the cognitive penalty of informational distraction from the physical penalty of energetic decay.

\section{Research Question and Hypothesis}

\textbf{Research Question}\\
Can the frame-by-frame phonetic entropy of a self-supervised acoustic model (wav2vec 2.0) function as an \textit{in silico} proxy for the RAMPHO buffer to objectively quantify and dissociate the cognitive penalties of Informational Masking (IM) versus Energetic Masking (EM)?

\vspace{1em}
\noindent \textbf{Hypothesis}\\
Phonetic entropy ($H$) will accurately track the RAMPHO implicit processing tipping point.
\begin{itemize}
    \item At high SNRs, an intelligible distractor will elicit significantly higher entropy (the IM penalty) than a phase-decorrelated distractor, reflecting semantic competition.
    \item At low SNRs, the decorrelated distractor will elicit the highest entropy (the EM penalty) by physically smearing the temporal fine structure, preventing the acoustic glimpsing necessary for target resolution.
\end{itemize}

\section{Methods}

\textbf{3.1 Stimuli Generation and Masker Conditions}\\
To decouple energetic and informational load, an English target narrative was mixed with three distinct maskers across a Signal-to-Noise Ratio (SNR) sweep (0, 5, 10, 15, and 20 dB, plus a 100 dB pristine baseline). Active speech levels were calibrated using ITU-T P.56 Active RMS normalization \citep{itu1993}.

The three masker conditions were defined as follows:
\begin{itemize}
    \item \textbf{Native Intelligible Masker (ENG):} A competing male English speaker. This condition introduces both fluctuating Energetic Masking and high Informational Masking via semantic competition.
    \item \textbf{Concentration Shield (Decorrelated Masker - CS):} The same English speaker, subjected to a targeted Digital Signal Processing (DSP) intervention. The speech-critical band (1--4 kHz) was isolated via Fast Fourier Transform (FFT), and the phase information was fully randomized. To prevent adversarial temporal smearing artifacts caused by hard spectral discontinuities, Hann window tapers were applied at the 1 kHz and 4 kHz crossover boundaries. This spectral surgery destroys the linguistic payload or Speech Transmission Index \citep{steeneken1980} while preserving the global RMS energy and low-frequency ($< 1$ kHz) spatial anchoring cues. This condition introduces zero IM but high continuous EM.
    \item \textbf{Speech-Shaped Noise (SSN):} A steady-state noise matched to the long-term average speech spectrum, serving as the standard control for pure Energetic Masking without temporal amplitude modulations.
\end{itemize}

\vspace{1em}
\noindent \textbf{3.2 The \textit{In Silico} RAMPHO Proxy}\\
To simulate the rapid, implicit phonological processing of the RAMPHO buffer, the audio mixtures were processed through a pre-trained self-supervised acoustic model (\texttt{facebook/wav2vec2-base-960h}) \citep{baevski2020}. To access sub-lexical representations, the Connectionist Temporal Classification (CTC) linear projection head was utilized. While wav2vec 2.0 employs a bidirectional transformer that utilizes non-causal future context---fundamentally exceeding human real-time processing constraints---it serves here as an idealized upper-bound proxy for implicit feature extraction before strict chronological constraints are engineered into future architectures. 

Audio was resampled to 16 kHz and normalized. For every 20 ms frame, the network executed a forward pass to extract the raw logits, followed by a Softmax activation to generate a probability distribution across the English vocabulary.

\vspace{1em}
\noindent \textbf{3.3 Metric Computation: Phonetic Entropy}\\
Implicit cognitive load was mathematically defined as the ambiguity of the phonetic categorization at any given frame. This was quantified by calculating the Shannon Entropy ($H$) of the Softmax probability distributions.

To ensure that a low entropy state genuinely reflects phonological binding rather than acoustic silence, the CTC ``blank'' token was explicitly excluded. The remaining probabilities were re-normalized over the active linguistic vocabulary ($K=31$):

\begin{equation}
H[n] = - \sum_{i=1}^{31} \left( \frac{P(x_i)}{1 - P(blank)} \right) \log_2 \left( \frac{P(x_i)}{1 - P(blank)} + \epsilon \right)
\end{equation}

Where $P(x_i)$ is the probability of phoneme/character $i$ at frame $n$, and $\epsilon$ is a negligible constant to prevent logarithmic evaluation of zero.

A low entropy value indicates that the network (proxying the RAMPHO buffer) confidently resolved a phonetic category, simulating automatic, effortless processing. A high entropy value indicates a flat probability distribution (phonetic ambiguity), simulating a failure of the implicit buffer and the requisite recruitment of explicit Working Memory to repair the stream.

\section{Results}

\textbf{Dissociation of Informational and Energetic Masking}\\
The frame-by-frame phonetic entropy ($H$) extracted from the \textit{in silico} RAMPHO proxy successfully captured a non-linear crossover between informational and energetic processing loads across the SNR sweep.

\vspace{1em}
\noindent \textbf{Release from Informational Masking (High SNR Regime)}\\
At favorable signal-to-noise ratios (15 to 20 dB), the simulation objectively quantified a semantic processing penalty. The Intelligible Native Masker (ENG) maintained a significantly elevated entropy state ($H \approx 0.16$ at 20 dB) compared to the steady-state Energetic Control (SSN; $H \approx 0.11$). Because peripheral audibility of the target is near-perfect at +20 dB, this gap represents pure Informational Masking: the central semantic competition driving ambiguity within the categorization layer.

When the distractor was subjected to spectral surgery via the Concentration Shield (CS), the entropy curve converged perfectly with the SSN baseline. By decorrelating the phase in the speech-critical band (1--4 kHz), the linguistic payload was destroyed. The network instantly discarded the stream as unparseable noise, providing a complete release from informational masking and returning the system to a low-effort state of automatic processing.

\vspace{1em}
\noindent \textbf{The Temporal Smearing Penalty (Low SNR Regime)}\\
At severe noise levels (0 to 5 dB), the relative processing costs inverted. The Intelligible Masker ($H \approx 0.67$ at 0 dB) outperformed the CS condition ($H \approx 0.85$), despite carrying a higher semantic payload.

This demonstrates the critical boundary of temporal glimpsing. The unmodified English masker, while semantically distracting, contains natural amplitude modulations that allow the network to extract target acoustic features during temporal dips. The phase-randomization of the CS condition destroyed these temporal gaps, transforming the distractor into a dense, adversarial energetic masker. The resulting spike in phonetic ambiguity empirically maps the exact threshold where the energetic penalty of structural smearing outweighs the cognitive relief of semantic destruction.

\begin{figure}[h]
\centering
\includegraphics[width=15cm]{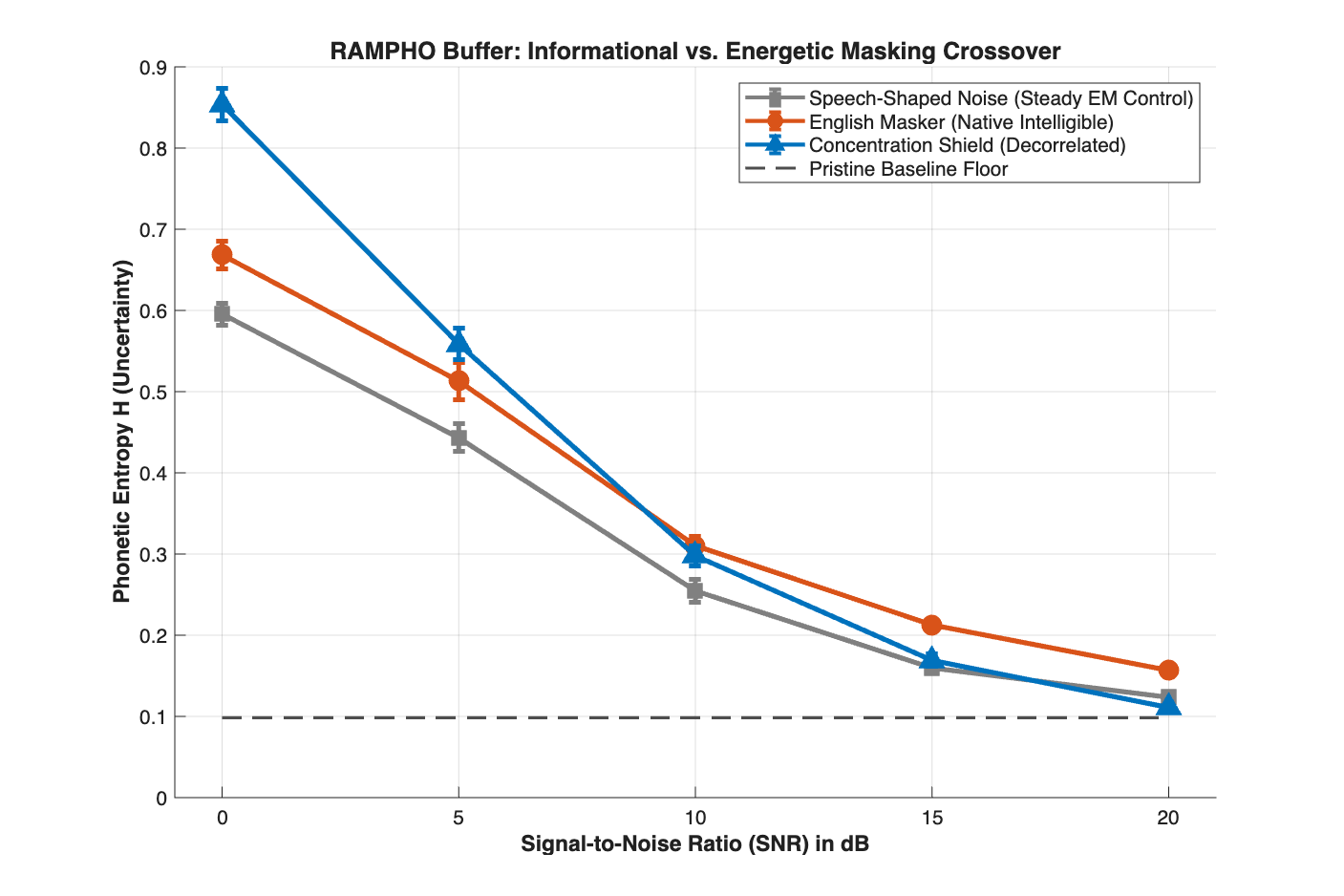}
\caption{RAMPHO Buffer: Informational vs. Energetic Masking Crossover. Frame-by-frame phonetic entropy ($H$) across the SNR sweep, demonstrating the objective cognitive-acoustic threshold.}
\label{fig:crossover}
\end{figure}

\section{Discussion \& Future Directions}

\textbf{The Cognitive-Acoustic Pareto Optimization}\\
These findings challenge the prevailing paradigm of strictly maximizing objective SNR in digital signal processing for hearing devices. The \textit{in silico} simulation proves that cognitive load is a zero-sum trade-off at the limits of audibility. Suppressing a distractor's semantic track (reducing IM) inherently degrades temporal landmarks (increasing EM). Therefore, next-generation ``cognitive hearing aids'' cannot rely on static spatial filters; they must dynamically solve a Cognitive-Acoustic Pareto Optimization problem, shifting their DSP strategy based on the immediate environmental SNR and the user's specific Working Memory capacity.

\vspace{1em}
\noindent \textbf{Architecting the ``Cognitive ASR''}\\
While phonetic entropy successfully proxies implicit RAMPHO buffer ambiguity, current self-supervised acoustic models (such as wav2vec 2.0) possess near-infinite context windows, failing to model the explicit memory constraints of the human listener. To evolve this pipeline into a rigorous \textit{in silico} testbed for clinical DSP algorithms, future architectures must intentionally cripple the ASR to simulate biological boundaries:

\begin{enumerate}
    \item \textbf{Simulating Echoic Memory Decay:} Biological sensory traces degrade rapidly (2--4 seconds). Future models will introduce an exponential temporal decay penalty ($\tau$) on the Transformer's self-attention matrix, forcing the model to lose historical acoustic context identically to human sensory decay.
    \item \textbf{Simulating Working Memory Capacity via Beam Search:} The ELU model dictates that explicit working memory is recruited to repair semantic streams when the RAMPHO buffer fails. This capacity limit will be simulated by aggressively constraining the Beam Search Decoder width (the number of parallel linguistic hypotheses maintained). By continuously sliding the target and masker against each other frame-by-frame, we will map the exact temporal overlap ($\Delta t$) that floods the constrained decoder with semantic competitors, forcing the target tracking to catastrophically collapse.
\end{enumerate}

\end{document}